\pdfoutput=1
\documentclass{article}
\usepackage{graphicx} 
\usepackage[utf8]{inputenc}
\usepackage[T1]{fontenc}
\usepackage[english]{babel}
\usepackage{lipsum}
\usepackage{textcomp}
\usepackage{fancyhdr}
\usepackage{amsmath,amssymb, amsthm}
\usepackage{lmodern}
\usepackage[a4paper]{geometry}
\usepackage{graphicx}
\usepackage{caption}
\usepackage{subcaption}
\usepackage{xcolor}
\usepackage{soul}
\usepackage{microtype}
\usepackage{hyperref}
\usepackage{enumitem} 
\usepackage{csquotes} 
\usepackage{siunitx} 
\usepackage{mathtools} 
\usepackage{algorithm}
\usepackage{algpseudocode}
\usepackage{siunitx}
\usepackage{booktabs}
\hypersetup{pdfstartview=XYZ}
\DeclareMathOperator{\wMAPE}{wMAPE}
\DeclareMathOperator{\MAPE}{MAPE}
\DeclareMathOperator{\MAE}{MAE}

\title{Deep Learning Approaches for Dynamic Mechanical Analysis of Viscoelastic Fiber Composites}

\author{Victor Hoffmann$^{1}$, 
Ilias Nahmed$^{1}$, 
Parisa Rastin$^{1,2}$ \\ 
Guénaël Cabanes$^{3}$, 
Julien Boisse$^{4}$\\
\textsuperscript{\rm 1}ENSMN,~~~ 
\textsuperscript{\rm 2}LORIA, UMR 7503,~~~  
\textsuperscript{\rm 3}LIPN, UMR 7030,~~~  
\textsuperscript{\rm 4}LEMTA, UMR 7563\\
\texttt{\{victor.hoffmann9,ilias.nahmed9\}@etu.univ-lorraine.fr}\\
\texttt{parisa.rastin@loria.fr} \\
\texttt{guenael.cabanes@lipn.univ-paris13.fr}\\
\texttt{julien.boisse@univ-lorraine.fr} 
}
\date{}

\begin{document}

\maketitle

\begin{abstract}
	The increased adoption of reinforced polymer (RP) composite materials, driven by eco-design standards, calls for a fine balance between lightness, stiffness, and effective vibration control. These materials are integral to enhancing comfort, safety, and energy efficiency. Dynamic Mechanical Analysis (DMA) characterizes viscoelastic behavior, yet there's a growing interest in using Machine Learning (ML) to expedite the design and understanding of microstructures. In this paper we aim to map microstructures to their mechanical properties using deep neural networks, speeding up the process and allowing for the generation of microstructures from desired properties.
\end{abstract}

\section{Introduction}

Designing strong, lightweight polymer-based composites with accurate mechanical behavior requires micromechanical approaches to describe a heterogeneous material as a set of constituents whose morphology, size distribution, and local mechanical behavior affect global behavior. Under dynamic loading, their viscoelastic properties, which are responsible for damping, must be taken into account during the design phase, as they can promote vibration isolation \cite{treviso_damping_2015}.\\

Numerous micromechanical codes are used to calculate the mechanical response to various loads applied to Representative Volume Elements (RVEs) of composite materials. Among these codes, Finite Element (FE) or Fast Fourier Transform (FFT) methods are widely used to compute local stress and strain fields as a function of time, frequency, or temperature under various loading conditions and to average them across RVE's for comparison with experimental observations \cite{MICHEL1999109}. \\

In the context of high performance composite design, Dynamic Mechanical Analysis (DMA) is one of the best ways to study and characterize the viscoelastic behavior of polymer matrix composites. DMA is known to be a privileged tool for the study of materials (especially polymers and rubbers) whose rheological behavior is viscoelastic by nature, i.e. involving irreversible dissipation of mechanical energy into heat. In its frequency mode DMA is based on harmonic steady-state strain excitations applied at $\omega$ pulsations to a material sample and recording the corresponding output stress signal. Complex algebra from experimental data gives access to the conservative (storage / glassy) or dissipative (loss / rubbery) modulus or compliance, which are the real and imaginary parts of their complex nature: $M^* (\omega)=M'(\omega)+jM''(\omega)$ for the modulus. This allows a full dynamic characterization of the material \cite{agbossou_modelling_1993}.\\

Characterizing the mechanical properties of numerically modeled composites is a complex challenge, usually requiring numerous calculations and considerable computational time when using conventional Finite Element Method (FEM) or Fast Fourier Transform (FFT) based codes \cite{Prakash_2009, MICHEL1999109}. With the advent of deep learning methods, a new approach to solving these issues has arisen, especially Convolutional Neural Networks (CNNs). However, any deep learning method still require a training database consisting of RVEs and their corresponding mechanical responses. \\

\begin{figure}
    \centering
    \includegraphics[width=1.\textwidth]{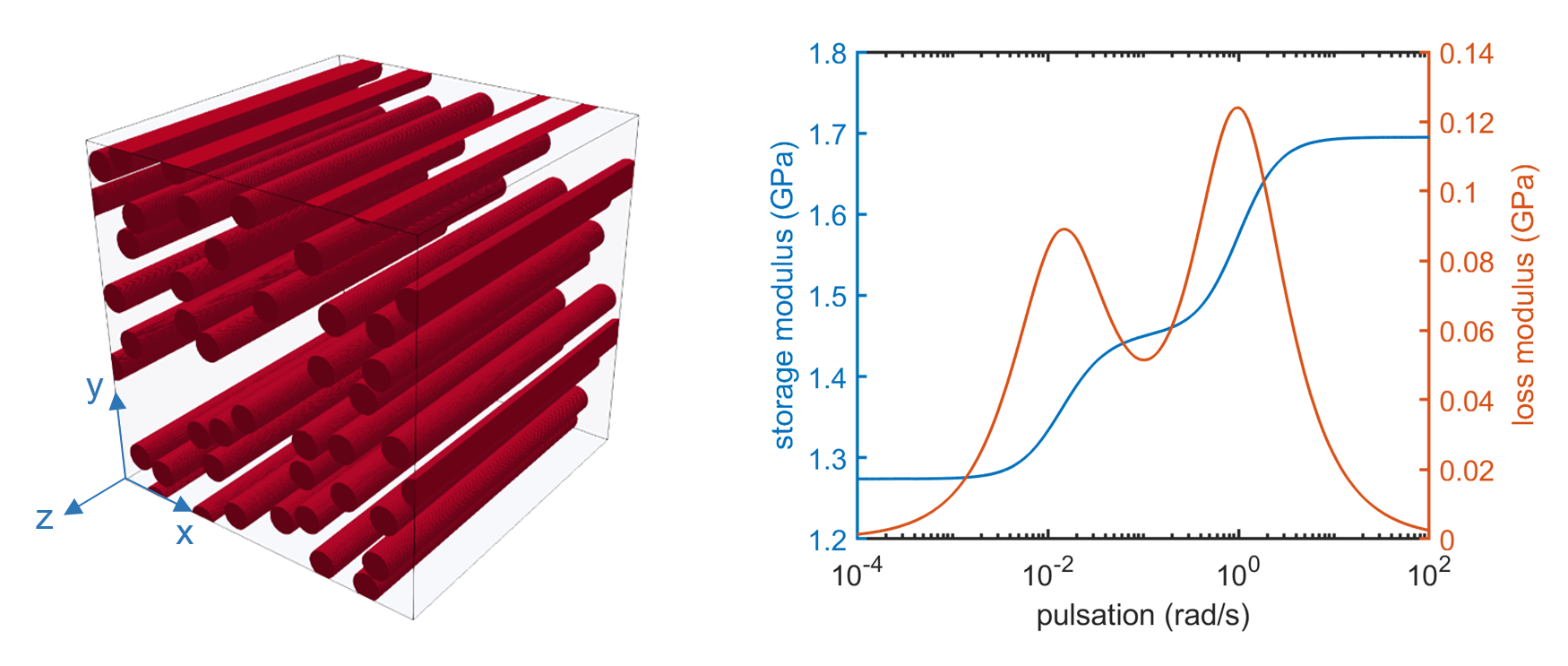}
    \caption{Modeled composite and corresponding storage and loss shear moduli. left: Representative Volume Element (RVE) of a 2D organized fiber reinforced composite with volume fractions 20\% fibers (in red) - 80\% matrix (transparent) satisfying periodic boundary conditions (PBC). Right: Storage $G_{xy}'$ (blue) and loss $G_{xy}''$ (red) shear moduli as a function of pulsation $\omega$ obtained from the computed DMA response of the RVE to a harmonic average shear strain load $\varepsilon_{xy}=\varepsilon_{0}cos(\omega t)$.}
    \label{fig:microstructure_dma_results}
\end{figure}

In this study, we focus on a first generated database consisting of RVEs of simple 2D organized fiber-reinforced composite microstructures built from a home-grown Matlab code, and their associated mechanical DMA signatures (see Fig. \ref{fig:microstructure_dma_results}) computed with the FFT-based code CraFT-Virtual-DMA \cite{jtcam:7456, labos_2022}. We present the testing of a convolutional neural network (CNN) algorithm trained on this first database.

\section{State of the Art}
\label{state_of_the_art}

Obtaining physical properties of simulated composite materials is a complex challenge that requires numerous calculations and a significant amount of computational time when using classical methods such as finite elements (FEM) or Fast Fourier Transform (FFT) methods. The emergence of deep learning methods, particularly Convolutional Neural Networks (CNNs), has provided a new approach to addressing these issues. A Convolutional Neural Network (CNN) \cite{OSheaN15} is a type of artificial neural network specifically designed for processing structured grid data, such as images and videos. CNNs are particularly well-suited for tasks involving visual recognition and understanding. They have become the foundation of wide range of applications such as Image Classification \cite{Guo2017}, Medical Imaging \cite{kayalibay2017cnnbased}, Environmental Monitoring \cite{Environmental}, Self-driving Cars \cite{Nugraha2017}, Natural Language Processing (NLP) \cite{Li2018} and many other. Over the last few years, CNNs have demonstrated notable efficiency in tackling these problems.\\

In 2018, Zijiang Yang \& al. \cite{yang_deep_2018} implemented 3D CNNs for the first time to model elastic homogenization linkages of 3D images of composite materials. The deep learning model introduced in the paper is utilized to understand the nonlinear relationship between the three-dimensional microstructure of the material and its overall stiffness. The research showcases the capability of this comprehensive approach to predict the stiffness of high-contrast elastic composites, considering a diverse array of microstructures. This model showed remarkable accuracy while requiring minimal computational resources for new assessments.

In 2020, Yixing Wang \& al. \cite{Wang_2020} introduce a data-driven and deep learning model to establish a part of the relationship between the structure and properties of polymer nanocomposites.  This model enables the exploration of structures and the identification of qualitative relationships between microstructure descriptors and mechanical properties. The authors introduce a novel deep learning method that combines convolutional neural networks with multi-task learning to create quantitative correlations between microstructures and property values.

Other techniques have been implemented to improve or simplify the CNN models used. For instance, A. Mann and S. Kalidindi \cite{mann_development_2022} worked on a novel CNN architecture with significantly fewer trainable parameters than current benchmarks. It achieves this by avoiding fully connected layers and utilizing 2-point spatial correlations of microstructures as input. This approach enhances robustness and enables cost-effective property closures by exploring the space of valid 2-point spatial correlations, which is a convex hull. It allows the generation of new valid 2-point spatial correlations from existing sets through convex combinations. This work highlights the advantages of using 2-point spatial correlations over voxelated discrete microstructures in current benchmarks.\\

Numerous papers have been published in this context \cite{xu_learning_2021, kim_exploration_2021, li_transfer_2018, xu_use_2022}, however, as far as we are aware, none have sought to predict the complete modulus curve encompassing both storage and loss moduli. Consequently, there is no established benchmark for comparing our model, except for the conventional methods employed in mechanical simulations.\\

In the next section, we will present the dataset used to validate our algorithm and introduce the proposed algorithm.

\section{Experimentation}

\subsection{Dataset}


The input data consists of \num{15000} distinct configurations of representative volume elements (RVEs) for unidirectional fiber-reinforced composite materials. All RVEs adhere to periodic boundary conditions. The fibers all have the same radius and vary from \num{1} to \num{150} per RVE, with volume fractions ranging from $0.05$ ($5$\%) to $0.75$ ($75$\%). A total of \num{100} random configurations have been simulated for each volume fraction (with a step of $0.005$) \\

\begin{figure}
    \centering
    \includegraphics[width=.3\textwidth]{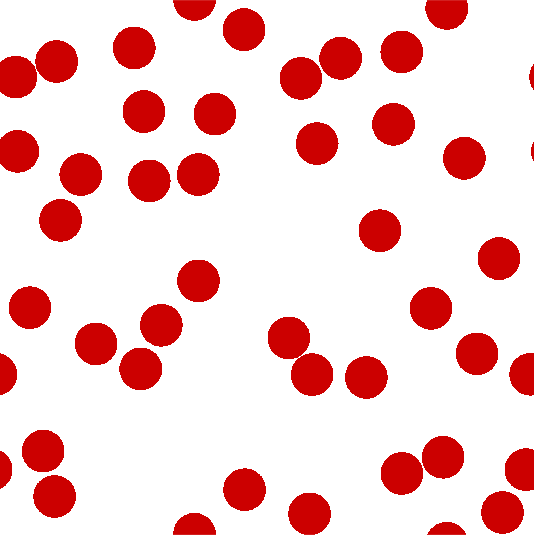}
    \includegraphics[width=.3\textwidth]{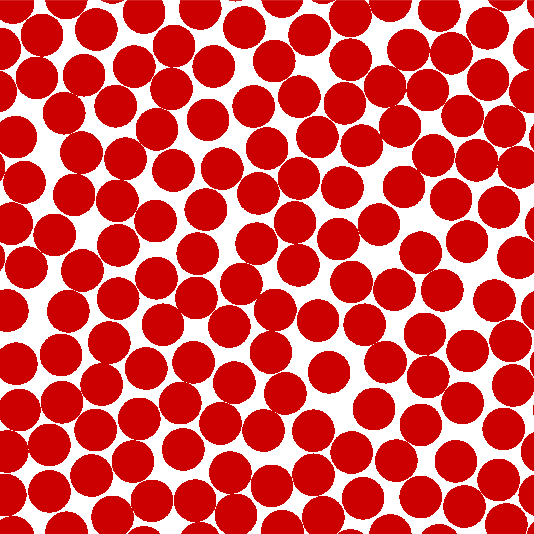}
    \includegraphics[width=.33\textwidth]{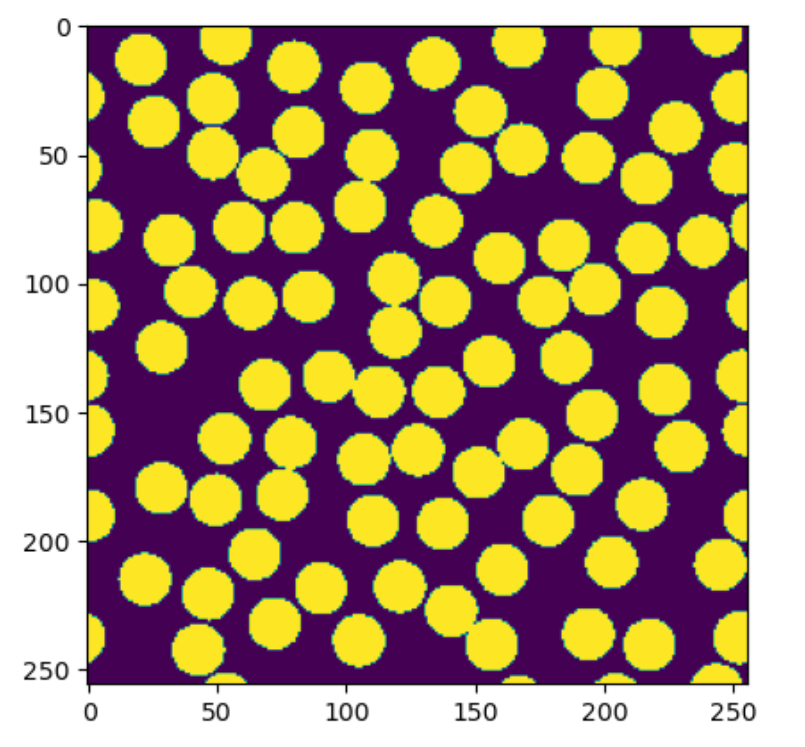}
    \caption{Examples of 2D Representative Volume Elements (RVEs) with volume fraction of 20\%, 70\% and 50\% fibers (from left to right). RVEs are depicted as $256\times256$ matrices (as shown on the right), where each element takes on a value of either $0.5$ (filler) or $-0.5$ (matrix). All RVEs respect periodic boundary conditions (PBC).}
    \label{fig:microstructure_image}
\end{figure}

Typically, RVEs are represented in three-dimensional matrices, but since we are studying unidirectional / 2D organized fiber materials, a cross-section of each matrix is sufficient. As a result, they will be represented by two-dimensional matrices of size $256\times256$, where each element is either $0.5$ (filler) or $-0.5$ (matrix). Examples of RVE representations are given in figure \ref{fig:microstructure_image}. \\

Each component (matrix and fibers) follows an isotropic viscoelastic behavior decribed by a 3D Standart Linear Solid (SLS) model with one relaxation time. The local constitutive law, {\it i.e.} stress as a function of strain, of each voxel is computed with the following complex linear relation in the frequency (or pulsation) domain
\begin{equation} \label{constitutive}
    \boldsymbol{\sigma}(\mathbf{x},j\omega) = \left[K(\mathbf{x},j\omega)-\frac{2}{3}G(\mathbf{x},j\omega)\right] \mathrm{Tr}\left[\boldsymbol{\varepsilon}(\mathbf{x},j\omega)\right] \mathbf{I} + 2G(\mathbf{x},j\omega)\boldsymbol{\varepsilon}(\mathbf{x},j\omega)
\end{equation}

where $\boldsymbol{\sigma}(\mathbf{x},j\omega)$ and $\boldsymbol{\varepsilon}(\mathbf{x},j\omega)$ are local stress and strain respectively as a function of position $\textbf{x}$ and pulsation $\omega$. $K(\mathbf{x},j\omega)$ and $G(\mathbf{x},j\omega)$ are local viscoelastic bulk and shear moduli which are computed as follow 
\begin{equation} \label{moduli}
    M(\mathbf{x},j\omega) = \frac{M_\infty (\mathbf{x}) + j\omega\tau_M (\mathbf{x}) M_0(\mathbf{x})}{1+j\omega\tau_M (\mathbf{x})} \quad \text{with} \quad M = K, G
\end{equation}

The local material mechanical parameters $M_\infty (\mathbf{x})$, $M_0 (\mathbf{x})$ and $\tau_M (\mathbf{x})$ are relaxed modulus, unrelaxed modulus and relaxation time respectively. \\

\begin{table}
\begin{center}
\begin{tabular}{ c | c | c | c | c | c | c }
        & $K_0$ (GPa) & $K_\infty$ (GPa) & $\tau_K$ (s) & $G_0$ (GPa) & $G_\infty$ (GPa) & $\tau_G$ (s) \\ 
\hline
 matrix & 8.6 & 7.33 & \textbf{1.} & 0.55 & 0.47 & \textbf{1.} \\  
 fiber  & 40.5556 & 4.05556 & 10. & 30.4167 & 3.04167 & 10. \\
\hline
\hline
        & $E_0$ (GPa) & $E_\infty$ (GPa) & $-$ & $\nu_0$  & $\nu_\infty$ & $-$ \\ 
\hline
 matrix & \textbf{1.6156} & \textbf{1.3805} & $-$ & \textbf{0.4687} & \textbf{0.4687} & $-$ \\  
 fiber  & \textbf{73} & 7.3 & $-$ & \textbf{0.2} & \textbf{0.2} & $-$ \\

\end{tabular}
\caption{Mechanical parameters for matrix and fibers. For the matrix: unrelaxed Young modulus ($E_0$), relaxed Young modulus ($E_\infty$), poisson coefficient ($\nu_0$) and relaxation time $\tau_{K,G}$ have been chosen considering a typical polypropylen material. For the fibers: unrelaxed young modulus ($E_0$) and poisson coefficient ($\nu_0$) have been taken from typical glass mechanical properties \cite{jtcam:7456}. A ratio of 10 between unrelaxed ($0$) on Relaxed ($\infty$) moduli as well as a relaxation time of $10s$ have been chosen artificially. For both materials: Poisson coefficient between unrelaxed ($0$) and relaxed states ($\infty$) have been kept constant, and, bulk ($K$) and shear ($G$) moduli have been calculated starting from Young moduli and Poisson coefficients.}
\label{parameters}
\end{center}
\end{table}

For each RVE, a mechanical response to an applied mean shear loading $\varepsilon_{xy}=\varepsilon_{0}cos(\omega t)$ have been computed using the CraFT Virtual DMA code. Average of the stress responses over the RVEs as a function of the pulsations $\omega$ are obtained directly from the simulations. DMA curves, {\it i.e.} storage and loss shear moduli, are real and imaginary parts of the complex shear modulus $G_{xy}^*(\omega)$
\begin{equation} \label{computed_moduli}
    G_{xy}^*(\omega) = \frac{\left<\sigma_{xy}(\textbf{x},j\omega)\right>_{RVE}}{\varepsilon_0} = G_{xy}'(\omega) + j G_{xy}''(\omega)
\end{equation}

In the context of our study the Mechanical parameters given in table \ref{parameters} for each material have been chosen to enlighten the two distinctive contributions of the matrix and fibers to the global averaged behavior of RVEs characterized by the DMA curves (Fig. \ref{fig:microstructure_dma_results}).\\

The output data consists of a set of two dynamic mechanical analysis (DMA) curves for each RVE: the shear storage modulus and the shear loss modulus, both provided in \si{\giga\pascal} as a function of the given angular frequency in \si{\radian\per\second}. Each curve is represented by a collection of \num{30} data points provided on a logarithmic scale of angular frequency. The objective of our model is thus to predict these two curves solely based on the representations of the RVEs. It turns out that under shear stress, both the shear storage modulus and the shear loss modulus remain invariant under a \num{90}° rotation and horizontal and vertical symmetries of the RVE. \\

In order for the model to understand periodic boundary conditions, the data has been augmented with \num{4} random (\num{2} horizontal and \num{2} vertical) translations of each RVE. The computer's used RAM (\num{32} GB) did not allow for further data expansion. Regarding the invariance under \num{90}° rotation, horizontal, and vertical symmetry, we opted for an equivariant convolutional neural network \cite{e2cnn} instead of data augmentation (cf. \ref{EqCNN}). \num{70}\% of the data was used for training the model, \num{15}\% was used for model validation, and the remaining 15\% was used for model testing. The batch size chosen for our model is \num{40}.

\subsection{Equivariant Convolutional Neural Networks}
\label{EqCNN}
Convolutional Neural Networks (CNNs) have gained substantial attention in recent years due to their effective performance in image processing tasks. However, traditional CNNs are not equipped to deal with transformations or symmetries in the input data. This limitation prompted the development of Equivariant Steerable CNNs (ESCNNs) \cite{e2cnn}, a form of convolutional neural network designed to handle specific transformations, such as rotations and symmetries without the need to see all possible transformed versions of the input data during training. \\

Let's denote the input to a layer as $x$, the transformed version of this input as $T(x)$ where $T$ is the transformation (for instance, rotation), and let $f$ denote the layer's operation. Then, the layer is equivariant if there exists a transformation $T'$ such that for all inputs $x$:
\begin{equation}\label{equivariant eq.}
f(T(x)) = T'(f(x))
\end{equation}

(\ref{equivariant eq.}) means that applying the transformation to the input and then applying the layer operation is the same as applying the layer operation first and then transforming the output.
Such equivariances are achieved through the use of steerable filters. \\

Steerable filters are special kinds of filters whose responses can be computed for any rotated version from a limited set of basis filters. This property gives them the power to detect features irrespective of their orientation in the image, making the network more robust and efficient. Specifically, in the context of ESCNNs, the steerability of these filters helps in achieving rotational equivariance. \\

Therefore, ESCNNs have allowed us to make our model remain invariant under a \num{90}° rotation and horizontal and vertical symmetries of the RVE without data augmentation. However, this has a computational cost since for each equivariant convolutional block, we have \num{8} parallel convolutional blocks (1 for each unique combination of \num{90}° rotation and symmetry).

\subsection{Architecture of the model}

\begin{figure}
    \centering
    \includegraphics[width = \textwidth]{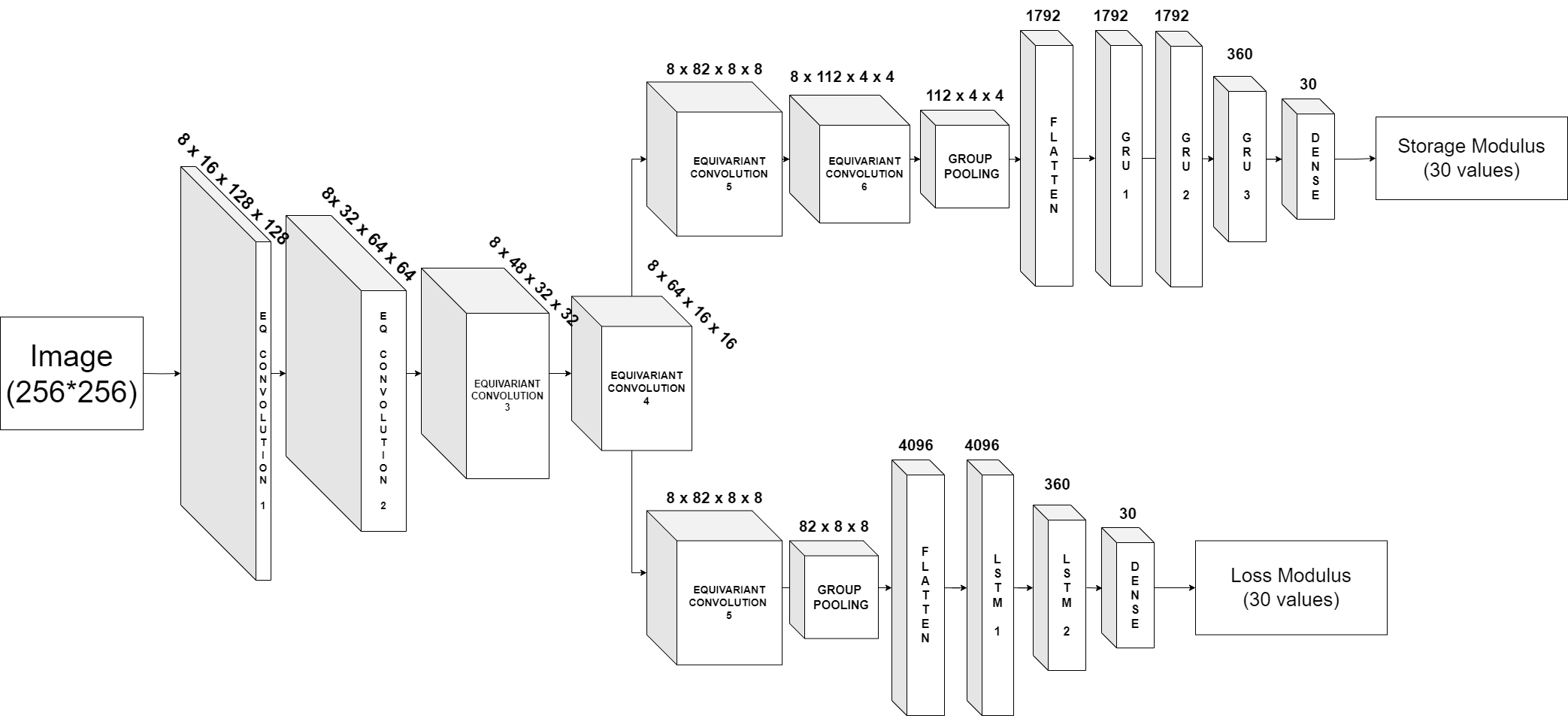}
    \caption{Architecture of the multi-task model. It consists of a common equivariant convolutional component, which divides itself into two branches: one for the shear storage modulus and another for the shear loss modulus. }
    \label{fig:EquiCNN_architecture}
\end{figure}

The figure \ref{fig:EquiCNN_architecture} shows a scheme of the architecture's model.

As our objective is to concurrently predict both the shear storage modulus and the shear loss modulus, we have opted for a multi-task model. This model consists of a common equivariant convolutional component, which subsequently diverges into two branches: one for the shear storage modulus and another for the shear loss modulus. Each branch is composed of at least one equivariant convolutional block, and several layers of Recurrent Neural Networks (RNNs). A Recurrent Neural Network (RNN) \cite{medsker2001recurrent}is a type of artificial neural network designed for sequential data processing. Despite their utility, basic RNNs have limitations, including the "vanishing gradient" problem, which makes it difficult to capture long-term dependencies. To address this, more advanced RNN variants, such as Long Short-Term Memory (LSTM) \cite{Graves2012} and Gated Recurrent Unit (GRU) \cite{Deygru}, have been developed. These variants are better equipped to capture and propagate information over longer sequences and are widely used in practice for various tasks. \\

Each convolutional block is composed of an equivariant convolutional layer with a kernel size of $3\times3$ and circular padding, a batch normalization layer to which we apply the ReLU activation function and an Antialiased MaxPooling layer. The purpose of the Antialiased MaxPooling is to apply MaxPooling to the kernel which has undergone a Gaussian filter that slightly blurs the kernel \cite{MaxPoolingAntialiased}. This enables the ESCNN to be invariant to slight translations of the image, allowing it, in conjunction with circular padding, to better comprehend the notions of periodic boundary conditions.

At the end of the convolutional blocks in both branches, a group pooling is applied. Group pooling aggregates the eight different convolutional blocks executed in parallel for each rotation/symmetry combination in order to extract features invariant to such transformations \cite{e2cnn}.\\

The two branches are not identical. The branch for the shear storage modulus comprises two equivariant convolutional blocks and three layers of Gated Recurrent Units (GRUs), between which we have applied a Dropout rate of 0.2. It turns out that the shear storage modulus is more sensitive to short-term dependencies and requires more feature extraction than the shear loss modulus. Conversely, the branch for the shear loss modulus has only one equivariant convolutional block and two layers of Long Short-Term Memory (LSTM) units, between which we have implemented a Dropout rate of 0.3. At the end of each RNN block in both branches, a Fully Connected layer is applied to obtain values in $\mathbb{R}$. Indeed, in Pytorch, the output of RNNs falls between 0 and 1, due to a sigmoid activation function that is implicitly applied \cite{PyTorch}.

\subsection{Performance metric}
The scale of both the shear storage modulus and the shear loss modulus varies according to the volume fraction of the microstructures given as input. The lower the volume fraction, the lower the minimum and maximum of each modulus. To avoid the model giving too much importance to error minimization on moduli of microstructures with high volume fractions and not enough importance to low volume fractions, a relative loss function is necessary. The Mean Absolute Percentage Error ($\MAPE$) is a classic choice for these types of situations. However, it poses problems for several loss modulus values that are very close to 0 for some low volume fraction cases, thus causing an explosion of this latter. To address this issue, we have opted for the weighted Mean Absolute Percentage Error ($\wMAPE$) which is defined for each batch and each modulus as:

\begin{equation} \label{wMAPE}
    \wMAPE = \frac{1}{N_{\text{seq.}}} \sum_{j \in \text{seq.}} \sum_{i \in \text{batch}} \frac{|y_{ij}-\hat{y}_{ij}|}{|y_{ij}|}
\end{equation}

where seq. stands for sequence, which is the discrete representation of the storage or loss modulus, $N_{\text{seq}} = 30$, $y_{ij}$ is the $j^{th}$ point of the $i^{th}$ sequence in the batch and $\hat{y_{ij}}$ is the prediction of $y_{ij}$. As shown in (\ref{wMAPE}) for each point of the moduli, we calculate the average absolute difference between the real and the predicted value of the batch, then we divide it by the average real value of the batch. Finally, we take the average batch error of each point of each modulus and sum the two error modulus. Therefore, the performance metric is equal to $\wMAPE_{\text{storage}} + \wMAPE_{\text{loss}}$. In the section \ref{results}, the Mean Absolute Error ($\MAE$) will also be provided for reference, which is defined for each batch and each modulus as:
\begin{equation}
    \MAE = \frac{1}{N_{\text{seq.}}} \sum_{j \in \text{seq.}} \frac{1}{N_{\text{batch}}} \sum_{i \in \text{batch}} |y_{ij} - \hat{y_{ij}}|
\end{equation}
with the same notations as in (\ref{wMAPE}) and $N_{\text{batch}} = 40$. \\

Regarding the optimizer, we have chosen Adam, with a learning rate of $0.001$ and an $L^2$ regularization term with $\alpha$ set to $0.0001$.

\section{Results} \label{results}

\begin{figure}[htbp]
  \centering
  \begin{minipage}{0.3\textwidth}  
    \centering
    \includegraphics[width=\textwidth, height = 6cm ]{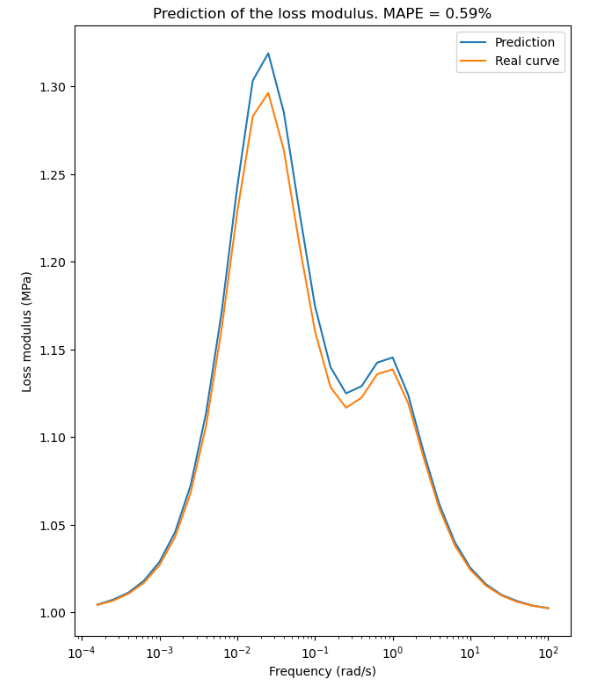}  
  \end{minipage}%
  \begin{minipage}{0.7\textwidth}  
    \centering
    \includegraphics[width=\textwidth, height = 6cm]{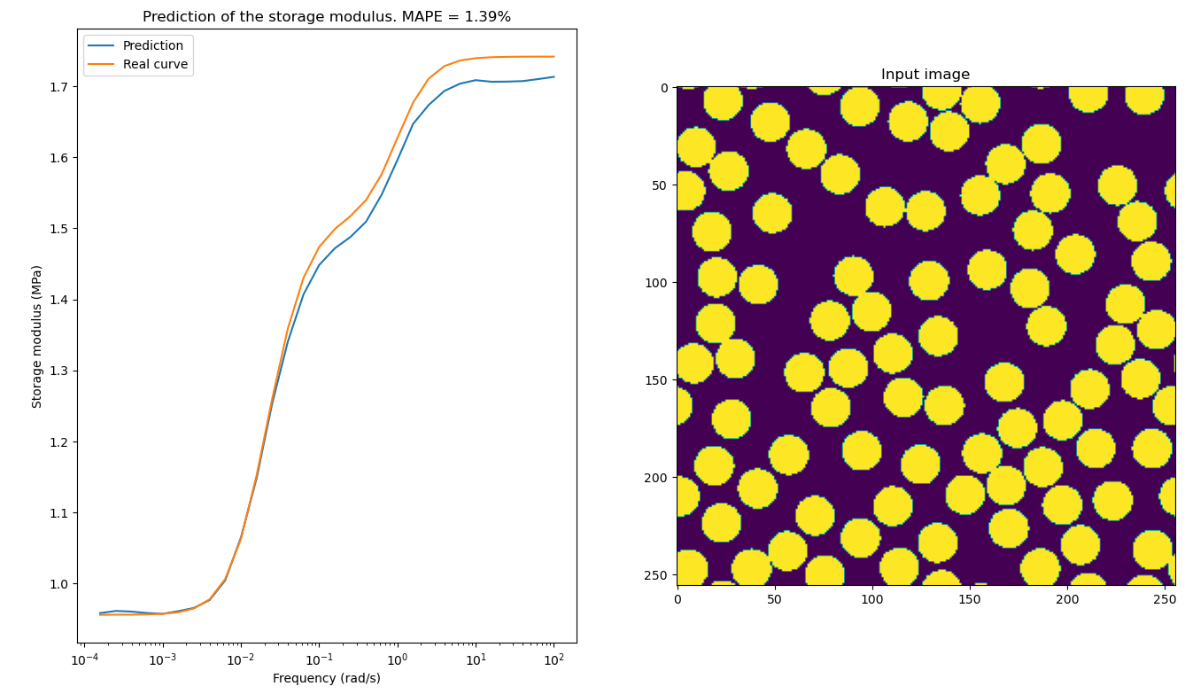}  
  \end{minipage}
    \caption{Prediction example of shear storage and loss moduli. Please take note that this concerns the Mean Absolute Percentage Error (MAPE), not the previously used weighted MAPE (wMAPE). Indeed, when there is just one prediction curve compared to a real curve (which is equivalent to a batch size set to 1), the wMAPE is equivalent to the MAPE.}
    \label{fig:prediction_example}
\end{figure}

The table \ref{results_table} shows the MAE and wMAPE of the model for both shear storage modulus and shear loss modulus. Each mean error also includes its standard deviation. As mentioned in \ref{state_of_the_art}, to the best of our knowledge, no prior literature exists that endeavors to predict the complete set of Dynamic Mechanical Analysis (DMA) curves for both shear storage modulus and shear loss modulus as functions of frequency. Consequently, we lack a benchmark against which to compare our findings. As depicted in Figure \ref{fig:prediction_example}, the challenge in predicting the loss modulus primarily lies within the range of intermediate frequencies, with values at the extremes being nearly negligible. Regarding the storage modulus, the difficulty of prediction often arises for high-frequency values.
\begin{table}[ht]
\begin{center}
\caption{wMAPE and MAE metrics of the model evaluated on the test set. The model has been trained on the wMAPE, the MAE is provided for reference.}
\begin{tabular}{lcc}
      \toprule
      & Shear Storage Modulus & Shear Loss Modulus \\
      \midrule
      wMAPE & 0.83\% +/- 0.13\% & 1.85\% +/- 0.25\% \\
      MAE & 0.0134 +/- 0.0027 & 0.0025 +/- 0.0005 \\
      \bottomrule

    \end{tabular}
    \label{results_table}
\end{center}
\end{table}

\begin{figure}
    \centering
    \includegraphics[scale = 0.7]{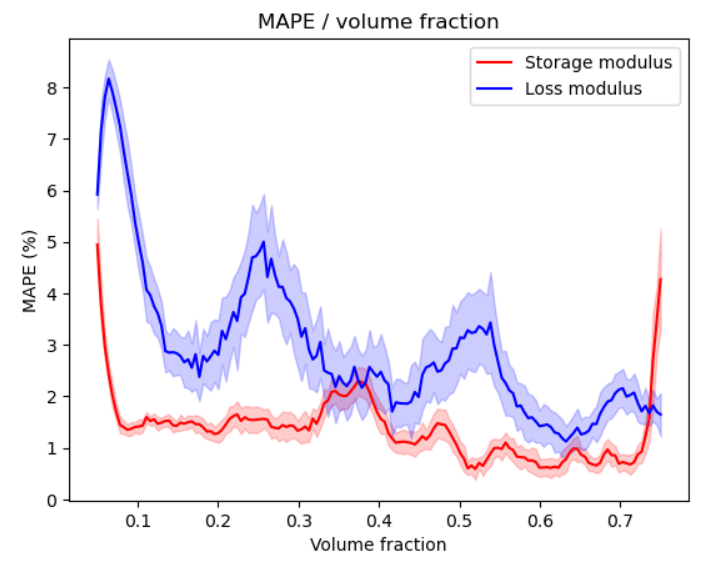}
    \caption{MAPE as a function of volume fraction. The average value of MAPE for each volume fraction has been plotted as a solid line, and the standard deviation of this value is depicted by the interval around the mean value. As mentioned in figure \ref{fig:prediction_example}, please take note that this concerns the Mean Absolute Percentage Error (MAPE), not the previously used weighted MAPE (wMAPE). To generate such curves, the model was assessed on each RVE separately, employing a batch size of 1, as opposed to the previous procedure.}
    \label{MAPE/fraction}
\end{figure}

With the purpose of assessing the robustness of our model, we conducted an analysis of its performance on each RVE as a function of the volume fraction. 
As depicted in Figure \ref{MAPE/fraction}, the model demonstrates robustness in predicting storage moduli for RVEs which volume fraction ranges from $0.07$ to $0.73$. However, the model's robustness appears to diminish when predicting loss moduli for RVEs with volume fractions near $0.3$ or $0.5$. Additionally, the uncertainty of the loss modulus is higher than the uncertainty of the storage modulus. The MAPE of the loss modulus for volume fractions between $0.05$ and $0.1$ is challenging to interpret, because the loss modulus exhibits values very close to $0$, which leads to a divergence in the MAPE values.

\bibliographystyle{IEEEtran}
\bibliography{main.bib}

\begin{thebibliography}{10}
\providecommand{\url}[1]{#1}
\csname url@samestyle\endcsname
\providecommand{\newblock}{\relax}
\providecommand{\bibinfo}[2]{#2}
\providecommand{\BIBentrySTDinterwordspacing}{\spaceskip=0pt\relax}
\providecommand{\BIBentryALTinterwordstretchfactor}{4}
\providecommand{\BIBentryALTinterwordspacing}{\spaceskip=\fontdimen2\font plus
\BIBentryALTinterwordstretchfactor\fontdimen3\font minus
  \fontdimen4\font\relax}
\providecommand{\BIBforeignlanguage}[2]{{%
\expandafter\ifx\csname l@#1\endcsname\relax
\typeout{** WARNING: IEEEtran.bst: No hyphenation pattern has been}%
\typeout{** loaded for the language `#1'. Using the pattern for}%
\typeout{** the default language instead.}%
\else
\language=\csname l@#1\endcsname
\fi
#2}}
\providecommand{\BIBdecl}{\relax}
\BIBdecl

\bibitem{treviso_damping_2015}
\BIBentryALTinterwordspacing
A.~Treviso, B.~Van~Genechten, D.~Mundo, and M.~Tournour, ``Damping in composite
  materials: Properties and models,'' \emph{Composites Part B: Engineering},
  vol.~78, pp. 144--152, 2015. [Online]. Available:
  \url{https://www.sciencedirect.com/science/article/pii/S1359836815002139}
\BIBentrySTDinterwordspacing

\bibitem{MICHEL1999109}
\BIBentryALTinterwordspacing
J.~Michel, H.~Moulinec, and P.~Suquet, ``Effective properties of composite
  materials with periodic microstructure: a computational approach,''
  \emph{Computer Methods in Applied Mechanics and Engineering}, vol. 172,
  no.~1, pp. 109--143, 1999. [Online]. Available:
  \url{https://www.sciencedirect.com/science/article/pii/S0045782598002278}
\BIBentrySTDinterwordspacing

\bibitem{agbossou_modelling_1993}
\BIBentryALTinterwordspacing
A.~Agbossou, A.~Bergeret, K.~Benzarti, and N.~Alberola, ``Modelling of the
  viscoelastic behaviour of amorphous thermoplastic/glass beads composites
  based on the evaluation of the complex poisson's ratio of the polymer
  matrix,'' \emph{Journal of Materials Science}, vol.~28, no.~7, pp.
  1963--1972, 1993. [Online]. Available:
  \url{https://doi.org/10.1007/BF00595770}
\BIBentrySTDinterwordspacing

\bibitem{Prakash_2009}
\BIBentryALTinterwordspacing
A.~Prakash and R.~A. Lebensohn, ``Simulation of micromechanical behavior of
  polycrystals: finite elements versus fast fourier transforms,''
  \emph{Modelling and Simulation in Materials Science and Engineering},
  vol.~17, no.~6, p. 064010, aug 2009. [Online]. Available:
  \url{https://dx.doi.org/10.1088/0965-0393/17/6/064010}
\BIBentrySTDinterwordspacing

\bibitem{jtcam:7456}
\BIBentryALTinterwordspacing
S.~André, J.~Boisse, and C.~Noûs, ``{An FFT solver used for virtual Dynamic
  Mechanical Analysis experiments: Application to a glassy/amorphous system and
  to a particulate composite},'' \emph{{Journal of Theoretical, Computational
  and Applied Mechanics}}, Apr. 2021. [Online]. Available:
  \url{http://jtcam.episciences.org/7456}
\BIBentrySTDinterwordspacing

\bibitem{labos_2022}
\BIBentryALTinterwordspacing
J.~Boisse and S.~André, ``{CraFT} virtual {DMA} · {GitLab}.'' [Online].
  Available:
  \url{https://gitlab.univ-lorraine.fr/labos/lemta/rheosol/craft-virtual-dma}
\BIBentrySTDinterwordspacing

\bibitem{OSheaN15}
\BIBentryALTinterwordspacing
K.~O'Shea and R.~Nash, ``An introduction to convolutional neural networks,''
  \emph{CoRR}, vol. abs/1511.08458, 2015. [Online]. Available:
  \url{http://arxiv.org/abs/1511.08458}
\BIBentrySTDinterwordspacing

\bibitem{Guo2017}
T.~Guo, J.~Dong, H.~Li, and Y.~Gao, ``Simple convolutional neural network on
  image classification,'' in \emph{2017 IEEE 2nd International Conference on
  Big Data Analysis (ICBDA)}, 2017, pp. 721--724.

\bibitem{kayalibay2017cnnbased}
B.~Kayalibay, G.~Jensen, and P.~van~der Smagt, ``Cnn-based segmentation of
  medical imaging data,'' 2017.

\bibitem{Environmental}
\BIBentryALTinterwordspacing
O.~İnik, ``Cnn hyper-parameter optimization for environmental sound
  classification,'' \emph{Applied Acoustics}, vol. 202, p. 109168, 2023.
  [Online]. Available:
  \url{https://www.sciencedirect.com/science/article/pii/S0003682X22005424}
\BIBentrySTDinterwordspacing

\bibitem{Nugraha2017}
B.~T. Nugraha, S.-F. Su, and Fahmizal, ``Towards self-driving car using
  convolutional neural network and road lane detector,'' in \emph{2017 2nd
  International Conference on Automation, Cognitive Science, Optics, Micro
  Electro-­Mechanical System, and Information Technology (ICACOMIT)}, 2017,
  pp. 65--69.

\bibitem{Li2018}
P.~Li, J.~Li, and G.~Wang, ``Application of convolutional neural network in
  natural language processing,'' in \emph{2018 15th International Computer
  Conference on Wavelet Active Media Technology and Information Processing
  (ICCWAMTIP)}, 2018, pp. 120--122.

\bibitem{yang_deep_2018}
\BIBentryALTinterwordspacing
Z.~Yang, Y.~C. Yabansu, R.~Al-Bahrani, W.-k. Liao, A.~N. Choudhary, S.~R.
  Kalidindi, and A.~Agrawal, ``\BIBforeignlanguage{en}{Deep learning approaches
  for mining structure-property linkages in high contrast composites from
  simulation datasets},'' \emph{\BIBforeignlanguage{en}{Computational Materials
  Science}}, vol. 151, pp. 278--287, Aug. 2018. [Online]. Available:
  \url{https://linkinghub.elsevier.com/retrieve/pii/S0927025618303215}
\BIBentrySTDinterwordspacing

\bibitem{Wang_2020}
\BIBentryALTinterwordspacing
Y.~Wang, M.~Zhang, A.~Lin, A.~Iyer, A.~S. Prasad, X.~Li, Y.~Zhang, L.~S.
  Schadler, W.~Chen, and L.~C. Brinson, ``Mining structure–property
  relationships in polymer nanocomposites using data driven finite element
  analysis and multi-task convolutional neural networks,'' \emph{Mol. Syst.
  Des. Eng.}, vol.~5, pp. 962--975, 2020. [Online]. Available:
  \url{http://dx.doi.org/10.1039/D0ME00020E}
\BIBentrySTDinterwordspacing

\bibitem{mann_development_2022}
\BIBentryALTinterwordspacing
A.~Mann and S.~R. Kalidindi, ``\BIBforeignlanguage{en}{Development of a
  {Robust} {CNN} {Model} for {Capturing} {Microstructure}-{Property} {Linkages}
  and {Building} {Property} {Closures} {Supporting} {Material} {Design}},''
  \emph{\BIBforeignlanguage{en}{Frontiers in Materials}}, vol.~9, p. 851085,
  Mar. 2022. [Online]. Available:
  \url{https://www.frontiersin.org/articles/10.3389/fmats.2022.851085/full}
\BIBentrySTDinterwordspacing

\bibitem{xu_learning_2021}
\BIBentryALTinterwordspacing
K.~Xu, A.~M. Tartakovsky, J.~Burghardt, and E.~Darve, ``Learning
  viscoelasticity models from indirect data using deep neural networks,''
  \emph{Computer Methods in Applied Mechanics and Engineering}, vol. 387, p.
  114124, 2021. [Online]. Available:
  \url{https://www.sciencedirect.com/science/article/pii/S0045782521004552}
\BIBentrySTDinterwordspacing

\bibitem{kim_exploration_2021}
\BIBentryALTinterwordspacing
Y.~Kim, H.~K. Park, J.~Jung, P.~Asghari-Rad, S.~Lee, J.~Y. Kim, H.~G. Jung, and
  H.~S. Kim, ``Exploration of optimal microstructure and mechanical properties
  in continuous microstructure space using a variational autoencoder,''
  \emph{Materials \& Design}, vol. 202, p. 109544, 2021. [Online]. Available:
  \url{https://linkinghub.elsevier.com/retrieve/pii/S0264127521000976}
\BIBentrySTDinterwordspacing

\bibitem{li_transfer_2018}
\BIBentryALTinterwordspacing
X.~Li, Y.~Zhang, H.~Zhao, C.~Burkhart, L.~C. Brinson, and W.~Chen, ``A transfer
  learning approach for microstructure reconstruction and structure-property
  predictions,'' \emph{Scientific Reports}, vol.~8, no.~1, p. 13461, 2018,
  bandiera\_abtest: a Cc\_license\_type: cc\_by Cg\_type: Nature Research
  Journals Number: 1 Primary\_atype: Research Publisher: Nature Publishing
  Group Subject\_term: Composites;Computational methods Subject\_term\_id:
  composites;computational-methods. [Online]. Available:
  \url{http://www.nature.com/articles/s41598-018-31571-7}
\BIBentrySTDinterwordspacing

\bibitem{xu_use_2022}
\BIBentryALTinterwordspacing
X.~Xu and N.~Gupta, ``Use of machine learning methods in syntactic foam
  design,'' in \emph{Encyclopedia of Materials: Plastics and Polymers},
  M.~S.~J. Hashmi, Ed.\hskip 1em plus 0.5em minus 0.4em\relax Elsevier, 2022,
  pp. 460--473. [Online]. Available:
  \url{https://www.sciencedirect.com/science/article/pii/B9780128203521001784}
\BIBentrySTDinterwordspacing

\bibitem{e2cnn}
M.~Weiler and G.~Cesa, ``{General E(2)-Equivariant Steerable CNNs},'' in
  \emph{Conference on Neural Information Processing Systems (NeurIPS)}, 2019.

\bibitem{medsker2001recurrent}
L.~R. Medsker and L.~Jain, ``Recurrent neural networks,'' \emph{Design and
  Applications}, vol.~5, no. 64-67, p.~2, 2001.

\bibitem{Graves2012}
\BIBentryALTinterwordspacing
A.~Graves, \emph{Long Short-Term Memory}.\hskip 1em plus 0.5em minus
  0.4em\relax Berlin, Heidelberg: Springer Berlin Heidelberg, 2012, pp. 37--45.
  [Online]. Available: \url{https://doi.org/10.1007/978-3-642-24797-2_4}
\BIBentrySTDinterwordspacing

\bibitem{Deygru}
R.~Dey and F.~M. Salem, ``Gate-variants of gated recurrent unit (gru) neural
  networks,'' in \emph{2017 IEEE 60th International Midwest Symposium on
  Circuits and Systems (MWSCAS)}, 2017, pp. 1597--1600.

\bibitem{MaxPoolingAntialiased}
\BIBentryALTinterwordspacing
R.~Zhang, ``Making convolutional networks shift-invariant again,'' \emph{CoRR},
  vol. abs/1904.11486, 2019. [Online]. Available:
  \url{http://arxiv.org/abs/1904.11486}
\BIBentrySTDinterwordspacing

\bibitem{PyTorch}
\BIBentryALTinterwordspacing
A.~Paszke, S.~Gross, F.~Massa, A.~Lerer, J.~Bradbury, G.~Chanan, T.~Killeen,
  Z.~Lin, N.~Gimelshein, L.~Antiga, A.~Desmaison, A.~K{\"{o}}pf, E.~Z. Yang,
  Z.~DeVito, M.~Raison, A.~Tejani, S.~Chilamkurthy, B.~Steiner, L.~Fang,
  J.~Bai, and S.~Chintala, ``Pytorch: An imperative style, high-performance
  deep learning library,'' \emph{CoRR}, vol. abs/1912.01703, 2019. [Online].
  Available: \url{http://arxiv.org/abs/1912.01703}
\BIBentrySTDinterwordspacing

\end{thebibliography}
\end{document}